\documentclass[letterpaper, 10 pt, conference]{ieeeconf}  

\IEEEoverridecommandlockouts                              

\usepackage{amsfonts}	
\usepackage{amsmath}	
\usepackage{amssymb}    
\usepackage{siunitx}
\usepackage{xfrac}    
\usepackage{pifont}   
\usepackage{dsfont}   

\usepackage{booktabs}
\usepackage{makecell}  
\usepackage[flushleft]{threeparttable}  
\usepackage{multirow}
\usepackage{rotating}
\usepackage{microtype}

\usepackage[dvipsnames]{xcolor}    
\let\labelindent\relax 
\definecolor{darkgreen}{rgb}{0.0,0.5,0.0}
\usepackage[inline]{enumitem} 
\usepackage{cite}  
\usepackage{xspace}    

\usepackage{array}
\usepackage{flushend}
\usepackage{float}

\renewcommand{\baselinestretch}{0.995}

\makeatletter
\let\NAT@parse\undefined
\makeatother

\usepackage[pdfencoding=auto, colorlinks=true]{hyperref} 
\usepackage[hang,flushmargin]{footmisc}

\usepackage[capitalize]{cleveref}
\crefname{section}{Sec.}{Secs.}
\Crefname{section}{Section}{Sections}
\Crefname{table}{Table}{Tables}
\crefname{table}{Tab.}{Tabs.}

\DeclareMathOperator*{\argmin}{arg\,min}

\newcommand{\method}{\mbox{ParkDiffusion}\xspace}
\newcommand{\jiarong}[1]{\textcolor{blue}{[Approved by Jiarong]}}
\title{\LARGE \bf
ParkDiffusion: Heterogeneous Multi-Agent Multi-Modal Trajectory Prediction for Automated Parking
using Diffusion Models}

\author{
Jiarong Wei$^{1,2}$,
Niclas Vödisch$^{1}$,
Anna Rehr$^{2}$,
Christian Feist$^{2}$,
and Abhinav Valada$^{1}$
\thanks{$^1$ Department of Computer Science, University of Freiburg, Germany.}%
\thanks{$^2$ CARIAD SE, Germany.}%
\thanks{A supplementary video is available at \href{https://youtu.be/tF3pd2DsOs4}{https://youtu.be/tF3pd2DsOs4}.}
\thanks{This work was partially funded by CARIAD and the German Research Foundation (DFG) Emmy Noether Program grant number 468878300.}
}

\begin{document}
\maketitle
\thispagestyle{empty}
\pagestyle{empty}


\begin{abstract}
    Automated parking is a critical feature of Advanced Driver Assistance Systems (ADAS), where accurate trajectory prediction is essential to bridge perception and planning modules.
Despite its significance, research in this domain remains relatively limited, with most existing studies concentrating on single-modal trajectory prediction of vehicles. 
In this work, we propose \method, a novel approach that predicts the trajectories of both vehicles and pedestrians in automated parking scenarios. \method employs diffusion models to capture the inherent uncertainty and multi-modality of future trajectories, incorporating several key innovations. 
First, we propose a dual map encoder that processes soft semantic cues and hard geometric constraints using a two-step cross-attention mechanism.
Second, we introduce an adaptive agent type embedding module, which dynamically conditions the prediction process on the distinct characteristics of vehicles and pedestrians.
Third, to ensure kinematic feasibility, our model outputs control signals that are subsequently used within a kinematic framework to generate physically feasible trajectories.
We evaluate \method on the Dragon Lake Parking (DLP) dataset and the Intersections Drone (inD) dataset.
Our work establishes a new baseline for heterogeneous trajectory prediction in parking scenarios, outperforming existing methods by a considerable margin.

\end{abstract}


\section{Introduction}

Automated parking has emerged as a prominent driver assistance system, driven by the increasing demand for convenience, safety, and space efficiency in urban environments. A critical component is trajectory prediction, which bridges the perception and planning modules and, hence, directly impacts the overall performance.
Nonetheless, most existing studies in automated parking discuss aspects such as perception~\cite{mu2023inverse, yang2021towards}, localization~\cite{liu2024less, kim2024visual}, and planning~\cite{dai2021long, zheng2024speeding}. 
Consequently, the vast majority of work toward trajectory prediction focuses on generic urban traffic~\cite{zhou2023query, zhang2024simpl,distelzweig2024entropy,distelzweig2024motion} or pedestrian-only scenarios~\cite{mao2023leapfrog, wong2024socialcircle, radwan2020multimodal} rather than the challenges that occur in parking environments.
\looseness=-1

\begin{figure}[t]
    \centering\includegraphics[width=\linewidth]{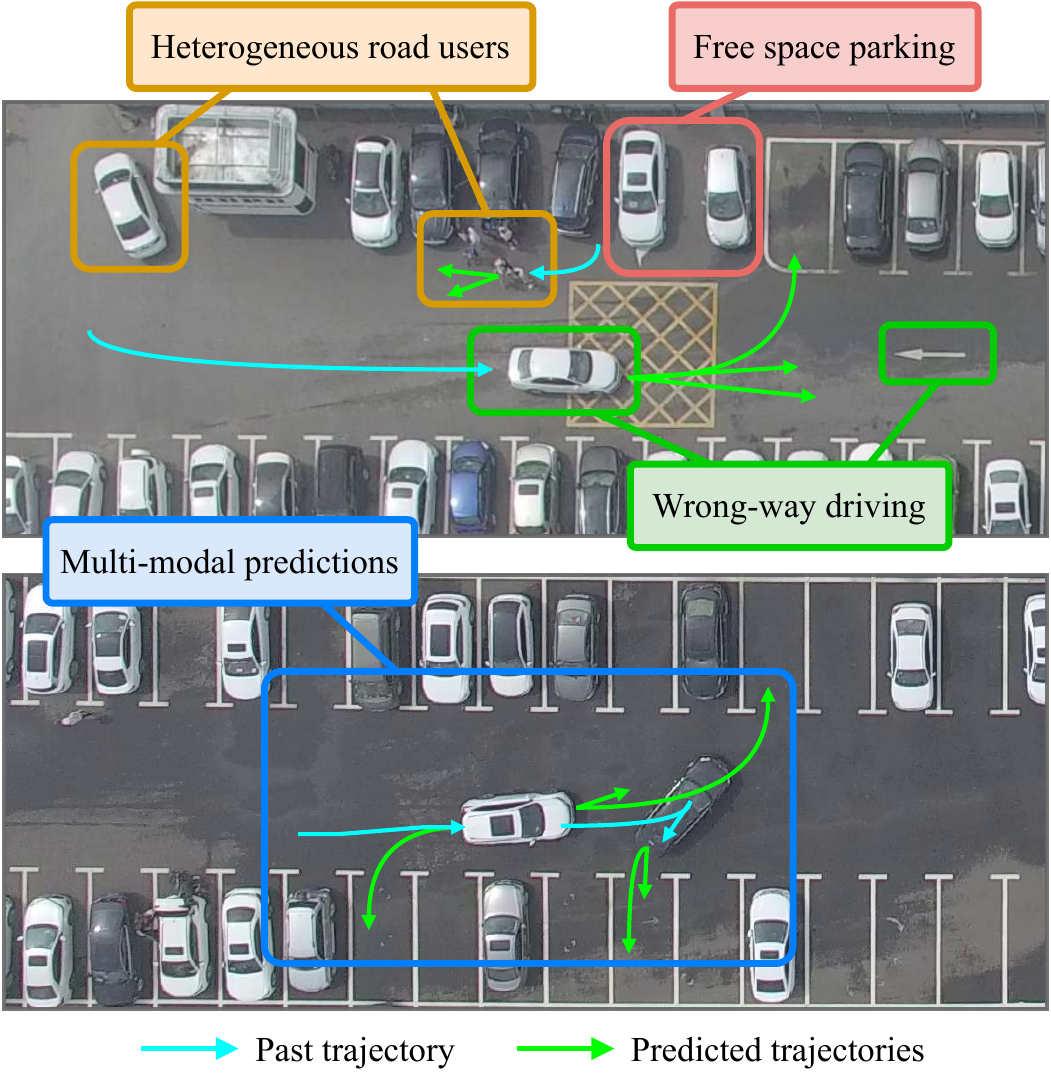}
    \vspace*{-.5cm}
    \caption{Trajectory prediction for automated parking faces several challenges. Road user behavior is often less constrained, with vehicles frequently parking outside designated spots or traveling in the wrong direction. Moreover, various road users often share the same space, especially in the absence of separate pedestrian pathways. Our proposed \method approach effectively addresses these problems by generating multi-modal trajectory predictions in scenarios with multiple heterogeneous agents.}
    \label{fig:cover}
    \vspace*{-.4cm}
\end{figure}

Unlike typical roadways, parking lots are often less structured and have fewer traffic regulations, as illustrated in \cref{fig:cover}.
They frequently lack defined opposing lanes, and vehicles may travel against one-way directions. 
Moreover, road user behavior is usually less predictable, with vehicles making abrupt turns into unoccupied parking spaces or reversing without adequate visibility. Finally, both vehicles and vulnerable road users (VRUs), such as pedestrians, often share the same space without dedicated sidewalks or pedestrian crossings. Although the driving speed in parking lots is typically lower than on standard roads, these challenges increase the risk of unsafe interactions.


In this work, we address trajectory prediction in the context of automated parking. Specifically, we propose the novel \method model, which addresses the following key aspects:
\begin{enumerate*}[label={(\arabic*)}]
    \item the varying behavior of heterogeneous road users, such as passenger cars and pedestrians;
    \item multi-agent forecasting, enabling the simultaneous prediction of future trajectories for multiple road users while accounting for their interactions; and
    \item multi-modal trajectories, facilitating the integration of probabilistic-driven approaches.
\end{enumerate*}

The objective of our work is to predict future trajectories of road users in parking scenarios, given their previous positions and additional contextual information. We build \method on the concept of diffusion~\cite{mao2023leapfrog} due to its capability of modeling complex probabilistic distributions. We propose a novel dual map encoder that separately processes soft traffic regulations such as lane markings and hard obstacles such as parked cars. We further include an agent type embedding to account for heterogeneous road users. Finally, we combine learning-based modules with a physics-driven approach that models the kinematics of different agent types.\looseness=-1

To summarize, the main contributions are as follows:
\begin{enumerate}[topsep=0pt]
    \item We introduce \method, a novel method for multi-agent and multi-modal trajectory prediction in challenging parking scenarios.
    \item We present the first work on heterogeneous trajectory prediction that explicitly considers VRUs in parking scenarios, enhancing the safety of automated parking.
    \item We integrate physics-based kinematic constraints to generate more realistic and feasible trajectories.
    \item We perform extensive experiments on the DLP~\cite{shen2022parkpredict+} and inD~\cite{bock2020ind} datasets which demonstrate that \method achieves state-of-the-art performance.
\end{enumerate}

\section{Related Work}

We present an overview of trajectory prediction, diffusion models for this task, and relevant works on automated parking. 


{\parskip=2pt
\noindent\textit{Trajectory Prediction:}
The task of trajectory prediction focuses on forecasting an agent’s motion based on past trajectories.
Related works evolve from traditional methods, such as the Extended Kalman Filter, to learning-based methods that offer a more comprehensive understanding of the environment.
SceneTransformer~\cite{ngiam2021scene} jointly predicts all agents’ behaviors, yielding consistent futures that account for their interactions.
MultiPath++~\cite{varadarajan2022multipath++} proposes a context-aware fusion of these elements and develops a multi-context gating fusion component for better feature representation.
QCNet~\cite{zhou2023query} introduces a query-centric paradigm for scene encoding, reusing past computations to improve prediction efficiency.
SIMPL~\cite{zhang2024simpl} proposes a global fusion module with symmetric directed message passing, enabling single-pass forecasting for all road users.
}

An important variation of the vanilla task accounts for heterogeneous agents, facilitating an advanced scene understanding by forecasting trajectories for different road users such as cars or pedestrians. 
Some methods utilize scene graph-structured models ~\cite{grimm2023heterogeneous, fang2023heterogeneous} to represent interactions among road users, incorporating both agent-to-agent and agent-to-environment interactions.
Another common approach is based on long short-term memory networks (LSTMs). 
TraPHic~\cite{chandra2019traphic} models inter-agent interactions with an LSTM-CNN hybrid network. TrafficPredict~\cite{ma2019trafficpredict} refines heterogeneous predictions using instance and category layers with an LSTM. 
In contrast to standard LSTM designs, Li~\textit{et~al.}~\cite{li2023real} employ hierarchical convolutional networks and multi-task learning to predict the trajectories of heterogeneous agents.
Other works employ attention-based approaches, e.g., Zhao~\textit{et~al.}~\cite{zhao2023trajectory} propose a dual attention model that captures interactions between road users and their impact on future trajectories. 
In our work, we account for heterogeneous agents via an agent type embedding module and by explicitly modeling the different kinematics of cars and vulnerable road users such as pedestrians. 
In parking scenarios, cars and pedestrians often share the same space, thereby increasing the risk of unsafe interactions and highlighting the need for type-specific considerations.


{\parskip=2pt
\noindent\textit{Diffusion Models for Trajectory Prediction:}
Diffusion-based generative models have emerged as powerful tools for capturing the inherent uncertainty and multi-modality of future motion. 
Recent studies explore diffusion models from various perspectives.
Liu~\textit{et~al.}~\cite{liu2024intention} decouple prediction uncertainty into intention and action components to enhance diffusion models’ capacity.
MotionDiffuser~\cite{jiang2023motiondiffuser} learns a joint multi-modal distribution for multi-agent trajectory prediction. 
Li~\textit{et~al.}~\cite{li2023multi} employ a coarse-to-fine approach, generating trajectory proposals using a conditional variational autoencoder, followed by refinement through a conditional diffusion model.
Lv~\textit{et~al.}~\cite{lv2024learning} propose an autoencoder diffusion model of pedestrian group relationships for multi-modal trajectory prediction.
Choi~\textit{et~al.}~\cite{choi2024dice} use diffusion models for efficient future trajectory prediction through an optimized sampling mechanism.
ControlTraj~\cite{zhu2024controltraj} integrates road network constraints to guide the geographical outcomes.
BCDiff~\cite{li2023bcdiff} predicts trajectories from instantaneous observations to address limited observation time.

One key drawback of diffusion models is their difficulty in real-time applications due to the large number of denoising steps required. To overcome this challenge, we leverage the Leapfrog Diffusion (LED) framework~\cite{mao2023leapfrog}, which enhances the diffusion model efficiency by introducing an LED Initializer that skips a large number of denoising steps. Although our approach is based on the Leapfrog skeleton, we introduce several key innovations, including the dual map encoder, agent type embedding, and kinematic refinement, as outlined in our list of contributions, resulting in significant performance improvements in both vehicles and pedestrians over the vanilla LED model.
To the best of our knowledge, this work is the first to utilize diffusion models for heterogeneous trajectory prediction in complex parking scenarios.
}


{\parskip=2pt
\noindent\textit{Automated Parking:}
Automated parking is a key component of driver assistance systems, driven by demand for enhanced comfort, safety, and spatial efficiency in dense urban areas.
Owing to challenges distinct from general urban driving, research on automated parking has examined components such as perception~\cite{mu2023inverse, yang2021towards}, planning~\cite{dai2021long, zheng2024speeding}, localization~\cite{liu2024less, kim2024visual}, SLAM~\cite{li2023avm, kang2021robust}, or combinations of these components~\cite{leu2022autonomous, li2024parkinge2e}. 
Nonetheless, despite posing a vital link between perception and planning, the task of trajectory prediction remains understudied.
Existing relevant works include ParkPredict~\cite{shen2020parkpredict} and ParkPredict++~\cite{shen2022parkpredict+}.
ParkPredict compares LSTM-based models with an Extended Kalman Filter, underscoring intent’s impact on future trajectories. 
ParkPredict++ extends this by leveraging convolutional and transformer-based architectures to extract spatiotemporal features from trajectory histories and bird’s-eye-view semantic maps.
However, both works adopt a sequential learning paradigm in which parking intent is inferred first, followed by the forecasting of future trajectories.
Additionally, these methods are limited to single-modal and vehicle-only trajectory prediction with low accuracy.
To the best of our knowledge, our work represents the first step toward trajectory prediction for multiple heterogeneous road users tailored for automated parking.
}
\begin{figure*}[t]
    \centering
    \includegraphics[width=\textwidth]{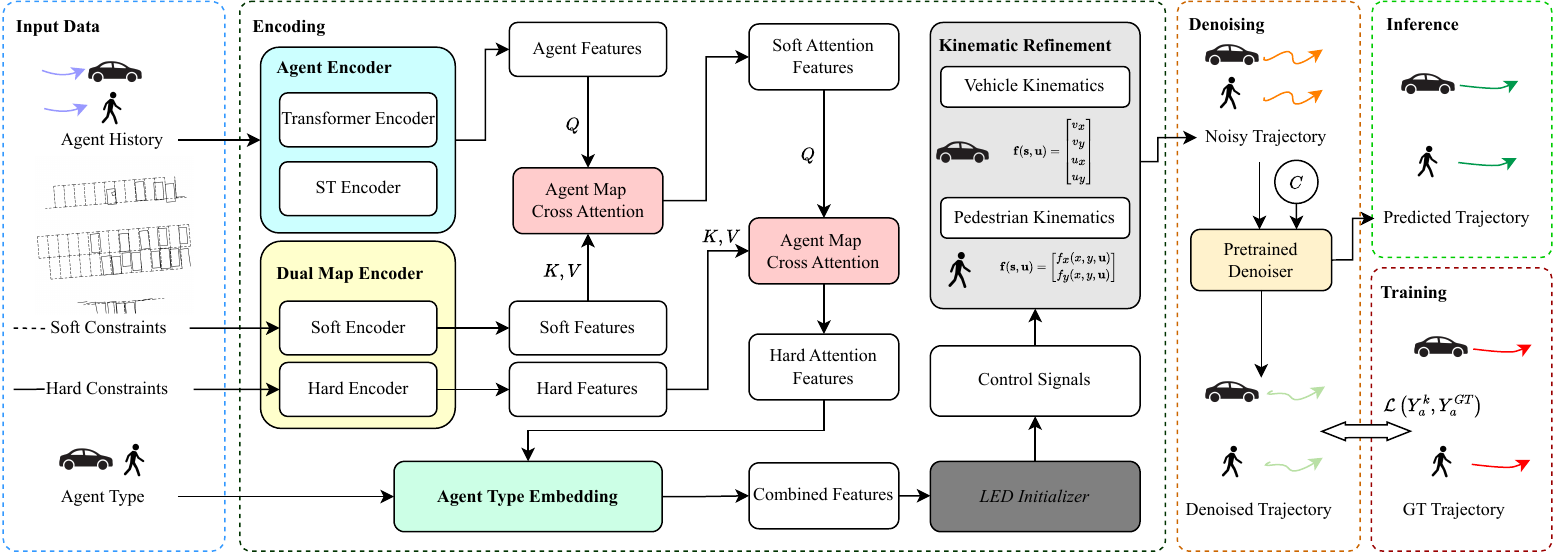}
    \vspace*{-.6cm}
    \caption{Overview of our \textbf{\method} approach for heterogeneous multi-agent multi-modal trajectory prediction. It takes as input agent histories, map polylines representing soft and hard constraints, and agent type information. The agent encoder and dual map encoder extract relevant features from the agents and maps, which are then fused through an attention mechanism. These fused features, along with the agent type data, are processed by the agent type embedding module. Next, the LED initializer~\cite{mao2023leapfrog} generates control signals that drive the kinematic refinement module to yield a noisy trajectory. Finally, the noisy trajectory, combined with contextual information, is passed through a pretrained denoiser. During training, the denoised output is compared with the ground truth, while during inference, the denoised trajectory serves as the final prediction.}
    \label{fig:overview}
    \vspace*{-.3cm}
\end{figure*}


\section{Method} 
In this section, we introduce our \method approach for heterogeneous multi-agent multi-modal trajectory prediction, as depicted in \cref{fig:overview}.
Following a formal definition of the research problem, we outline the key components, which include: 
\begin{enumerate*}[label={(\arabic*)}]
    \item an ego vehicle-centric data preprocessing pipeline; 
    \item an agent encoder that combines a transformer encoder with a spatiotemporal encoder; 
    \item a dual map encoder that separately processes soft and hard map constraints, followed by an attention-based feature fusion mechanism; 
    \item an agent type embedding module to facilitate agent-specific feature representation; and 
    \item a kinematic refinement module to generate more physically feasible trajectories.
\end{enumerate*}
We also introduce the denoiser and the training procedure of our method.\looseness=-1

\vspace*{-0.1cm}
\subsection{Problem Formulation}
The primary objective of our work is to predict future trajectories of all road users within the perception range of the ego vehicle, given their previous positions and additional context information. 
We denote the set of all agents as \(\mathcal{A} = \{a_\mathit{ego}, a_1, a_2, \dots, a_N\}\), comprising the ego vehicle \(a_\mathit{ego}\) and other road users including cars and pedestrians. 
Each agent \(a \in \mathcal{A}\) is observed for \(T_p\) past timestamps, and its motion is predicted for \(T_f\) future timestamps.
Let \(X_a = [x_a^{-T_p+1}, x_a^{-T_p+2}, \dots, x_a^{0}] \in \mathbb{R}^{T_p \times 2}\) denote the past trajectories of agent \(a\). 
Collectively, we have \(X_{\mathcal{A}} = \{X_{a_\mathit{ego}}, X_{a_1}, X_{a_2}, \dots, X_{a_N}\}\).
The context is given by a map representation \(M\) consisting of map polylines and static obstacle polylines.
Our goal is to predict for each agent a sequence of future positions \(Y_a = [y_a^{1}, y_a^{2}, \dots, y_a^{T_f}] \in \mathbb{R}^{T_f \times 2}\).
The multi-modal nature of the prediction is captured by defining \(\mathcal{Y}_a = \{Y_a^1, \dots, Y_a^K\}\) to yield \(K\) plausible trajectories, where \(Y_a^k \in \mathcal{Y}_a\) represents the \(k^{th}\) candidate trajectory of agent \(a\). 

We define our model as \(\mathcal{M}_\theta(\cdot)\) with parameter \(\theta\) such that:
\begin{equation}
    \mathcal{M}_\theta(X_{\mathcal{A}}, M) = P_\theta(\mathcal{Y}_{\mathcal{A}} \mid X_{\mathcal{A}}, M)
\end{equation}
\(P_\theta\) represents the conditional probability distribution over the future trajectories  \(\mathcal{Y}_{\mathcal{A}}\) given the past observations \(X_{\mathcal{A}}\) and the map context \(M\). The overall learning problem is:
\begin{equation}
    \theta^* = \argmin_{\theta} \sum_{a \in \mathcal{A}} \min_{k \in \{1,\dots,K\}} \mathcal{L} \left(Y_a^k, Y_a\right) \,,
\end{equation}
where \(\mathcal{L}\) measures the discrepancy between predicted trajectories and the ground truth to minimize the best-match error.\looseness=-1

\vspace*{-0.1cm}
\subsection{Data Preprocessing}

We first preprocess the data into ego vehicle-centric samples.
For each sample, we collect all agent and context information within a \qty{20}{\meter} radius of the ego vehicle and transform all relevant information into the ego vehicle coordinate system.
The state of each agent is represented as a 12-dimensional feature vector in the ego vehicle coordinate system: \((x, y, h, v, a_x, a_y, x^{\textit{rel}}, y^{\textit{rel}}, h^{\textit{rel}}, v^{\textit{rel}}, a_x^{\textit{rel}}, a_y^{\textit{rel}})\), where \(x\) and \(y\) denote the coordinates, \(h\) represents the heading angle, \(v\) is the velocity, \(a_x\) and \(a_y\) correspond to the acceleration components along the \(x\) and \(y\) axes, and the quantities with the superscript \(\textit{rel}\) are computed relative to the ego vehicle.
Importantly, the velocity of each agent is computed as a signed quantity, which is crucial in the context of automated parking with frequent reverse maneuvers.

\vspace*{-0.1cm}
\subsection{Encoding}

In this section, we introduce the key components for encoding, including the agent encoder, dual map encoder, agent type embedding, and kinematic refinement.

{\parskip=2pt
\noindent\textit{Agent Encoder:}
The agent encoder aims to capture the spatiotemporal features of the past trajectories of each agent.
This module comprises two complementary steps.
First, the agent history is processed via a linear projection followed by two transformer encoder layers with self-attention, yielding refined features with a residual connection. 
Additionally, a spatiotemporal encoder employs a convolution layer to extract local spatial patterns from the trajectory data, and a Gated Recurrent Unit (GRU) aggregates these features temporally to produce a feature embedding. 
The resulting agent features are then concatenated to serve as the basis for subsequent context fusion with map and agent type information in our diffusion framework, denoted by \(\mathbf{e}_a\).
}


{\parskip=2pt
\noindent\textit{Dual Map Encoder:}
To robustly capture the environmental context, we design a dual map encoder that separately processes map information and static objects, both represented as polylines. 
While map data provides soft constraints (e.g., lane markings that could be crossed), static objects pose hard constraints (e.g., vehicles parked in parking spots that need to be avoided).
We denote the corresponding sets of soft and hard polylines as \(\mathcal{P}_\text{soft} = \{p_1, p_2, \dots, p_{N_\text{soft}}\}\) and \(\mathcal{P}_\text{hard} = \{q_1, q_2, \dots, q_{N_\text{hard}}\}\), respectively.} 

Our dual map encoder consists of two parallel branches. 
The soft branch processes \(\mathcal{P}_\text{soft}\) using a two-layer Multi-Layer Perceptron (MLP) to generate map feature embeddings \(\mathbf{f}_\text{soft}\) that capture high-level map semantic features. 
Similarly, the hard branch processes \(\mathcal{P}_\text{hard}\) via a separate two-layer MLP to produce features \(\mathbf{f}_{\text{hard}}\) that emphasize geometric constraints. After encoding, both sets of polyline features are fused using a two-step cross-attention mechanism. First, the agent features are refined by attending to the soft polyline features:
\begin{equation}
    \mathbf{e}_{a, \text{soft}} = \operatorname{softmax}\!\left(\frac{\mathbf{e}_a\, \mathbf{f}_{\text{soft}}^\top}{\sqrt{d}}\right) \mathbf{f}_{\text{soft}} \,,
\end{equation}
where \(\mathbf{e}_a\) denotes the agent feature, \(d\) denotes the feature dimension, and \(\mathbf{f}_{\text{soft}}\) is the collection of soft constraints embeddings. 
Next, the refined agent feature further attends to the hard polyline features:
\begin{equation}
    \mathbf{e}_{a, \text{map}} = \mathbf{e}_{a, \text{soft}} + \operatorname{softmax}\!\left(\frac{\mathbf{e}_{a, \text{soft}}\, \mathbf{f}_{\text{hard}}^\top}{\sqrt{d}}\right) \mathbf{f}_{\text{hard}} \,,
\end{equation}
with \(\mathbf{f}_{\text{hard}}\) denoting the collection of hard embeddings. 
This residual connection integrates both sources of contextual information, yielding a map-conditioned agent representation.


{\parskip=2pt
\noindent\textit{Agent Type Embedding:}
To account for heterogeneous agents, we employ an agent type embedding mechanism that modulates the map-conditioned agent features \(\mathbf{e}_{a, \text{map}}\) based on their type. 
Each agent is associated with a type label, such as \textit{vehicle} or \textit{pedestrian}. 
Rather than using a simple lookup, the agent type information is first mapped to a learnable embedding and then processed through an MLP layer to generate modulation parameters \(\gamma\) and \(\beta\). 
For each agent, the combined feature \(\mathbf{f}_c\) is modulated as follows:
\begin{equation}
    \mathbf{f}_c = \mathbf{e}_{a, \text{map}} \cdot (1 + \gamma) + \beta \,,
\end{equation}
where the element-wise multiplication scales the features and the addition of \(\beta\) shifts them. 
This adaptive modulation enables the network to dynamically condition the agent features on their type, thereby improving the downstream trajectory prediction performance.
}


\subsection{Kinematic Refinement}

A potential problem with black-box models for motion forecasting is that model outputs can be physically infeasible or lack the general characteristics of natural motion~\cite{cui2020deep}~\cite{westny2024diffusion}. 
To address this problem, we employ kinematic constraint layers for both vehicles and pedestrians, bridging the transformation from control signals to the refined trajectories. 
For any agent \(a\), the position at timestamp \(t\) is defined as:
\begin{equation}
    \mathbf{p}_a(t) = (p_{a,x}(t),\, p_{a,y}(t)) \,,
\end{equation}
the velocity as:
\begin{equation}
    \mathbf{v}_a(t) = (v_{a,x}(t),\, v_{a,y}(t)) \,,
\end{equation}
and the control input as:
\begin{equation}
    \mathbf{u}_a(t) = (u_{a,x}(t),\, u_{a,y}(t)).
\end{equation}

Note that \( \mathbf{u}_a(t) \) is the output of the Leapfrog Diffusion Initializer~\cite{mao2023leapfrog}, as shown in \cref{fig:overview}, which takes the combined feature \( \mathbf{f}_c \) of the agent type embedding module as input. 
We also attach a probability branch (a simple MLP layer that takes the combined feature \( \mathbf{f}_c \) as input and outputs probabilities) inside the Initializer to learn the probabilities of the multi-modal trajectory predictions.
For further details on the LED Initializer, please refer to the work by Mao~\textit{et~al.}~\cite{mao2023leapfrog}.

{\parskip=3pt
\noindent\textit{Vehicle Kinematics:}
The state of vehicles is defined as:
\begin{equation}
    \mathbf{z}_a(t) = (\mathbf{p}_a(t),\, \mathbf{v}_a(t)) \,,
\end{equation}

The kinematics are modeled via a point-mass system with acceleration control inputs. The state evolves according to:
\begin{equation}
    \dot{\mathbf{p}}_a(t) = \mathbf{v}_a(t), \quad \dot{\mathbf{v}}_a(t) = \mathbf{u}_a(t) \,.
\end{equation}
The control input is constrained by
\begin{equation}
    \|\mathbf{u}_a(t)\| \leq \mu g \,,
\end{equation}
where \(\mu = 0.7\) is the road adhesion coefficient and \(g = \qty[per-mode=fraction]{9.81}{\meter\per\second^2}\) is the gravitational acceleration.
}


{\parskip=3pt
\noindent\textit{Pedestrian Kinematics:}
The state of pedestrians is represented solely by the position:
\begin{equation}
\mathbf{z}_a(t) = \mathbf{p}_a(t) \,.
\end{equation}
Since the motion patterns of pedestrians are generally more complex, we employ a first-order neural ordinary differential equation to model their kinematics. 
Their state evolves as:
\begin{equation}
    \dot{\mathbf{p}}_a(t) = f_a(\mathbf{p}_a(t), \mathbf{u}_a(t)) \,,
\end{equation}
where \(f_a\) is a neural network that parameterizes the state transition based on the control signal.
}


{\parskip=3pt
\noindent\textit{State Transitions:}
To integrate the kinematics, we employ Heun's method. 
Let \(\mathbf{F}_a(\mathbf{z}_a(t), \mathbf{u}_a(t))\) denote the time derivative of the state. First, an Euler prediction is computed:
\begin{equation}
    \mathbf{z}_a^{\text{Euler}}(t+\Delta t) = \mathbf{z}_a(t) + \Delta t\, \mathbf{F}_a(\mathbf{z}_a(t), \mathbf{u}_a(t)) \,.
\end{equation}
Then, a correction update is applied:
\begin{equation}
\begin{aligned}
    \mathbf{z}_a(t+\Delta t) &= \mathbf{z}_a(t) + \frac{\Delta t}{2}( \mathbf{F}_a(\mathbf{z}_a(t), \mathbf{u}_a(t)) \\
                             &\quad + \mathbf{F}_a(\mathbf{z}_a^{\text{Euler}}(t+\Delta t), \mathbf{u}_a(t)) )\,.
\end{aligned}
\end{equation}
By iteratively applying this integration for each future timestamp \(t=1, \dots, T_f\), we extract the position components from \(\mathbf{z}_a(t)\) to form the candidate trajectories, denoted as \(\tilde{Y}\). They are then refined by the denoiser module to remove the diffusion noise and produce the final predictions.
}

\vspace*{-0.1cm}
\subsection{Denoising}
The denoiser refines the candidate trajectories generated by the kinematic constraints. Given a candidate trajectory~\( \tilde{Y} \) obtained after kinematic refinement, Gaussian noise is first added according to a predefined diffusion schedule. 
The denoiser $\varepsilon_\theta$ then estimates this noise, conditioned on a context vector \(C\) derived from the agent feature encoding, the dual map encoder, and the agent type embedding. The reverse diffusion update is given by:
\begin{equation}
    \hat{Y} = \frac{1}{\sqrt{\alpha_\beta}} \left( \tilde{Y} - \frac{1-\alpha_\beta}{\sqrt{1-\bar{\alpha}_\beta}}\, \varepsilon_\theta(\tilde{Y},\beta,C) \right) + \sigma_\beta\, z,
\end{equation}
where \(\alpha_\beta\), \(\bar{\alpha}_\beta\), and \(\sigma_\beta\) are determined by the diffusion schedule and \(z\sim\mathcal{N}(0,I)\). This procedure progressively removes the injected noise, yielding the final refined trajectory \(\hat{Y}\), which serves as one candidate output \(Y_a^k\) for agent \(a\), forming part of the multi-modal prediction set \(\mathcal{Y}_a\).

\vspace*{-0.1cm}
\subsection{Training}
  
Our training procedure is divided into two stages. 
In the first stage, ground-truth future trajectories are perturbed with Gaussian noise according to a predefined diffusion schedule. The denoiser is pretrained to predict the injected noise by minimizing the mean squared error:
\begin{equation}
    \mathcal{L}_{\text{denoiser}} = \| \varepsilon - \varepsilon_\theta(\tilde{Y},\beta,C) \|^2 \,,
\end{equation}
where \(\varepsilon\) is the added noise, \(\tilde{Y}\) is the noisy trajectory obtained by perturbing the ground truth, \(\beta\) is the diffusion timestamp, and \(C\) is the conditioning context derived from the agent feature encoding, dual map encoder, and agent type embedding.
\looseness=-1

In the second stage, with the denoiser frozen, we train the encoding part (as shown in \cref{fig:overview}) using a weighted \(L_2\) reconstruction loss. 
We denote the ground-truth trajectory of an agent \(a\) to be \(Y_a^{GT} \in \mathbb{R}^{T_f \times 2}\), and the \(K\) candidate trajectories output by the denoiser as \(\mathcal{Y}_a = \{Y_a^1, \dots, Y_a^K\}\). The reconstruction loss is defined as:
\begin{equation}
    \mathcal{L}_{\text{diffuser}} = \frac{1}{|\mathcal{A}|} \sum_{a \in \mathcal{A}} \min_{k \in \{1,\dots,K\}} \| Y_a^k - Y_a^{GT} \|_W^2 \,,
\end{equation}
where \(\|\cdot\|_W^2\) denotes a weighted \(L_2\) norm. Additionally, the probability branch that yields candidate likelihoods for multi-modal prediction is supervised with a cross-entropy loss computed on the candidate probabilities. The loss targets are derived from the reconstruction errors, ensuring that the candidate with the lowest error is treated as the correct label.

\section{Experiments}

In this section, we first introduce the datasets and implementation details, then compare our proposed \method with several baselines, and finally present extensive ablation studies to analyze the impact of each component and evaluate the robustness of our approach.


\begin{table*}[t]
\centering
\caption{Trajectory Prediction Results on the Dragon Lake Parking Dataset}
\vspace*{-0.3cm}
\label{table:results-dlp}
\setlength\tabcolsep{4pt}
\begin{threeparttable}
    \begin{tabular}{l | ccc | ccc | ccc}
        \toprule
        & \multicolumn{3}{c|}{\textbf{Vehicle}} & \multicolumn{3}{c|}{\textbf{Pedestrian}} & \multicolumn{3}{c}{\textbf{All}} \\
        \textbf{Method} & minADE [\unit{\meter}] & minFDE [\unit{\meter}] & MR [\%] & minADE [\unit{\meter}] & minFDE [\unit{\meter}] & MR [\%] & minADE [\unit{\meter}] & minFDE [\unit{\meter}] & MR [\%] \\
        \midrule
        EKF 
        & 1.30 & 2.90 & 53.0
        & 0.53 & 1.10 & 15.0
        & 0.75 & 1.60 & 25.8 \\
        ParkPredict+~\cite{shen2022parkpredict+}
        & 0.80 & 2.10 & 46.0 
        & -- & -- & -- 
        & -- & -- & -- \\
        MultiPath++~\cite{varadarajan2022multipath++}
        & 0.93 & 1.75 & 31.0 
        & 0.60 & 0.92 & 7.5
        & 0.67 & 1.10 & 13.0 \\
        SceneTransformer~\cite{ngiam2021scene} 
        & 0.85 & 1.64 & 28.0
        & 0.92 & 1.29 & 26.4
        & 0.90 & 1.39 & 26.9 \\
        QCNet~\cite{zhou2023query}
        & 0.33 & 0.57 & 1.1
        & \underline{0.44} & \underline{0.71} & \underline{2.5}
        & \underline{0.41} & \underline{0.67} & \underline{2.1} \\
        SIMPL~\cite{zhang2024simpl}  
        & \underline{0.29} & \underline{0.36} & \underline{0.3} 
        & 0.87 & 0.88 & 10.6
        & 0.71 & 0.73 & 7.7 \\
        ParkDiffusion (\textit{ours})
        & \textbf{0.17} & \textbf{0.24} & \textbf{0.2} 
        & \textbf{0.15} & \textbf{0.32} & \textbf{0.6} 
        & \textbf{0.16} & \textbf{0.30} & \textbf{0.5} \\
        \bottomrule
    \end{tabular}
    We compare \method with several baseline methods on the DLP dataset~\cite{shen2022parkpredict+} using a prediction horizon of \qty{4}{\second}. We report the mean across the \texttt{val} split of the minimum displacement error (minADE), the minimum final displacement error (minFDE), and the miss rate (MR). Bold and underlined values indicate the best and second-best scores, respectively. Our \method approach outperforms all baselines across the board by a large margin.
\end{threeparttable}
\end{table*}

\begin{table*}[t]
\centering
\caption{Trajectory Prediction Results on the Intersections Drone Dataset}
\vspace*{-0.3cm}
\label{table:results-ind}
\setlength\tabcolsep{4pt} 
\begin{threeparttable}
    \begin{tabular}{l | ccc | ccc | ccc}
        \toprule
        & \multicolumn{3}{c|}{\textbf{Vehicle}} & \multicolumn{3}{c|}{\textbf{Pedestrian}} & \multicolumn{3}{c}{\textbf{All}} \\
        \textbf{Method} & minADE [\unit{\meter}] & minFDE [\unit{\meter}] & MR [\%] & minADE [\unit{\meter}] & minFDE [\unit{\meter}] & MR [\%] & minADE [\unit{\meter}] & minFDE [\unit{\meter}] & MR [\%] \\
        \midrule
        EKF  
        & 3.08 & 6.73 & 78.9 
        & 0.66 & 1.34 & 22.7
        & 1.92 & 4.14 & 51.9 \\
        MultiPath++~\cite{varadarajan2022multipath++} 
        & 0.91 & 1.56 & 4.5
        & \underline{0.43} & 0.77 & 5.3
        & 0.68 & 1.18 & 15.2 \\
        SceneTransformer~\cite{ngiam2021scene} 
        & 0.75 &0.88 & 2.3 
        & 1.57 &1.77 & 32.2
        & 1.14 &1.31 & 9.7 \\
        QCNet~\cite{zhou2023query} 
        & 0.78 & 1.36 & 2.1
        & 0.73 & 1.19 & 13.0
        & 0.76 & 1.28 & \underline{7.3} \\
        SIMPL~\cite{zhang2024simpl} 
        & \textbf{0.46} & \underline{0.67} & \textbf{0.1}
        & 0.46 & \underline{0.49} & \underline{2.2}
        & \underline{0.46} & \underline{0.58} & \textbf{1.1} \\
        ParkDiffusion (\textit{ours}) 
        & \underline{0.58} & \textbf{0.51} & \underline{0.5}
        & \textbf{0.20} & \textbf{0.45} & \textbf{1.8}
        & \textbf{0.40} & \textbf{0.48} & \textbf{1.1} \\
        \bottomrule
    \end{tabular}
    We also compare our \method approach with several baseline methods on the inD dataset~\cite{bock2020ind}. We report the mean across the \texttt{val} split of the minimum displacement error (minADE), the minimum final displacement error (minFDE), and the miss rate (MR). Bold and underlined values indicate the best and second-best scores, respectively. Although SIMPL shows strong performance in some vehicle metrics, our \method achieves competitive or superior overall results, particularly excelling in pedestrian prediction accuracy. 
\end{threeparttable}
\end{table*}

\vspace*{-0.1cm}
\subsection{Evaluation Metrics}
We evaluate the performance of the model using standard trajectory prediction metrics. 
The minimum Average Displacement Error~(minADE) calculates the Euclidean distance in meters between the ground-truth trajectory and the best of \(K\) predicted trajectories as an average of all future time steps over all the agents.
On the other hand, the minimum Final Displacement Error~(minFDE) concerns the prediction error at the final time step, reflecting long-term performance.
The Miss Rate~(MR) is defined as the percentage of agents whose final prediction error exceeds \qty{2.0}{\meter}.


\begin{figure*}[t]
    \centering
    \begin{minipage}[b]{0.24\textwidth}
        \centering
        \includegraphics[width=\linewidth]{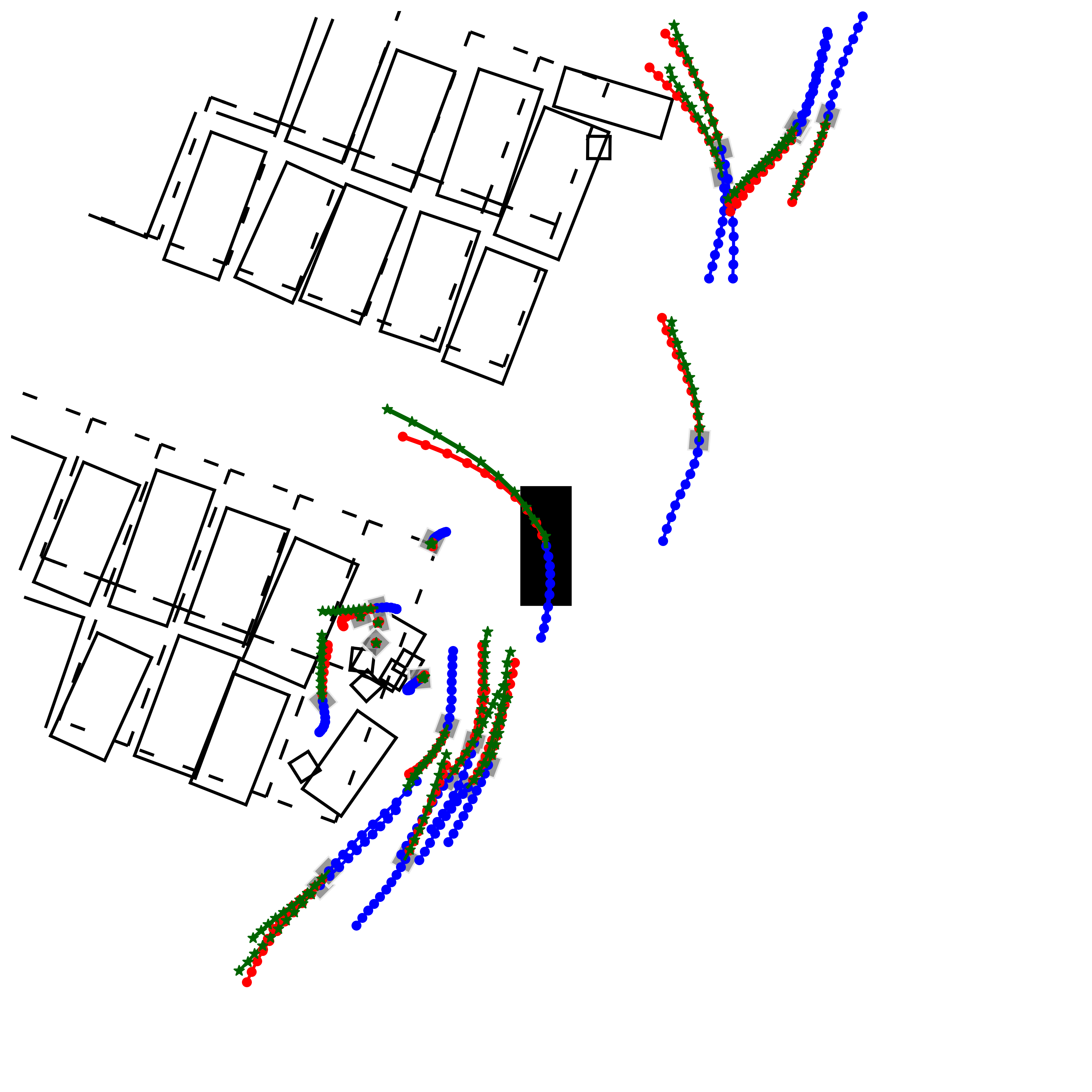}\\[0.3em]
    \end{minipage}
    \hfill
    \begin{minipage}[b]{0.24\textwidth}
        \centering
        \includegraphics[width=\linewidth]{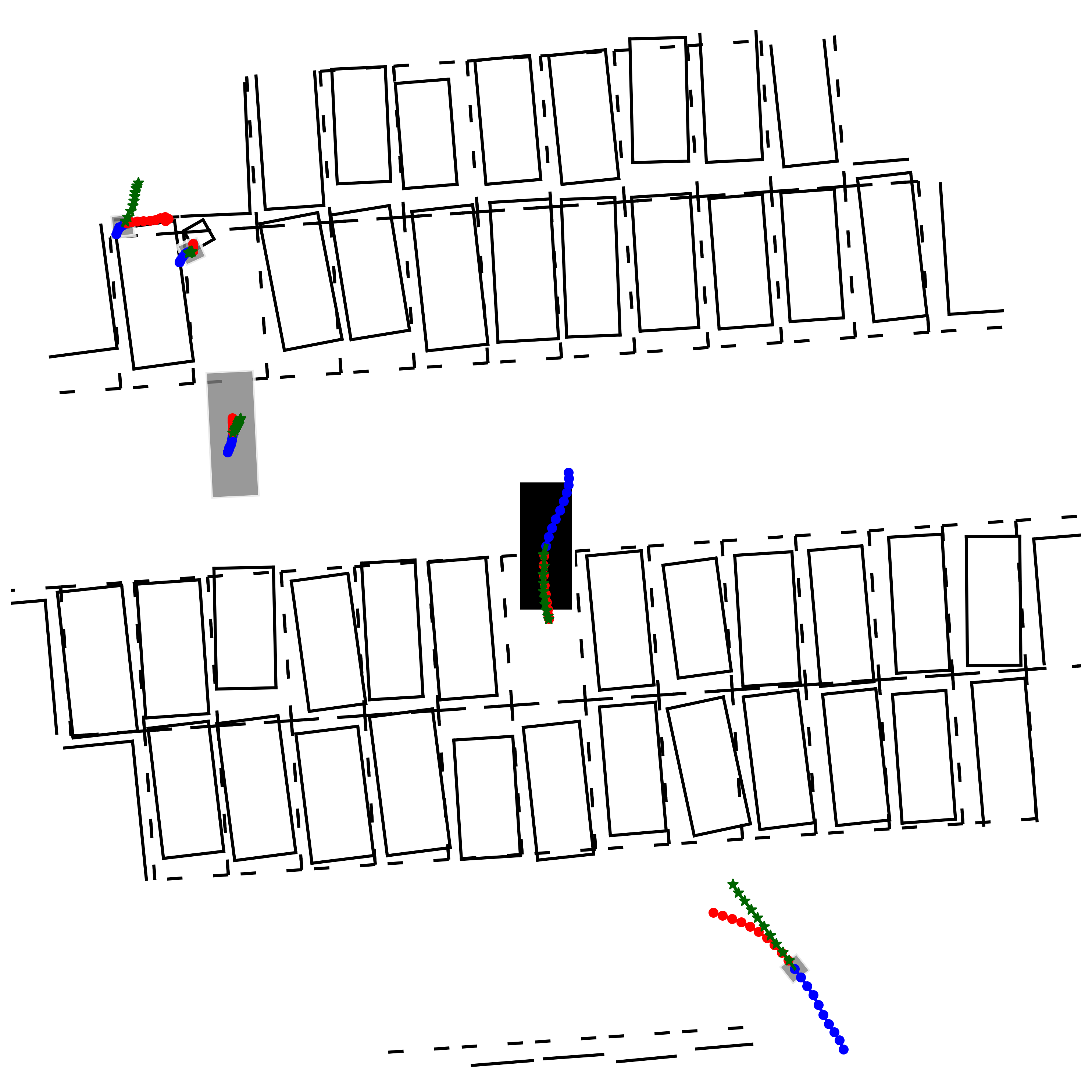}\\[0.3em]
    \end{minipage}
    \hfill
    \begin{minipage}[b]{0.24\textwidth}
        \centering
        \includegraphics[width=\linewidth]{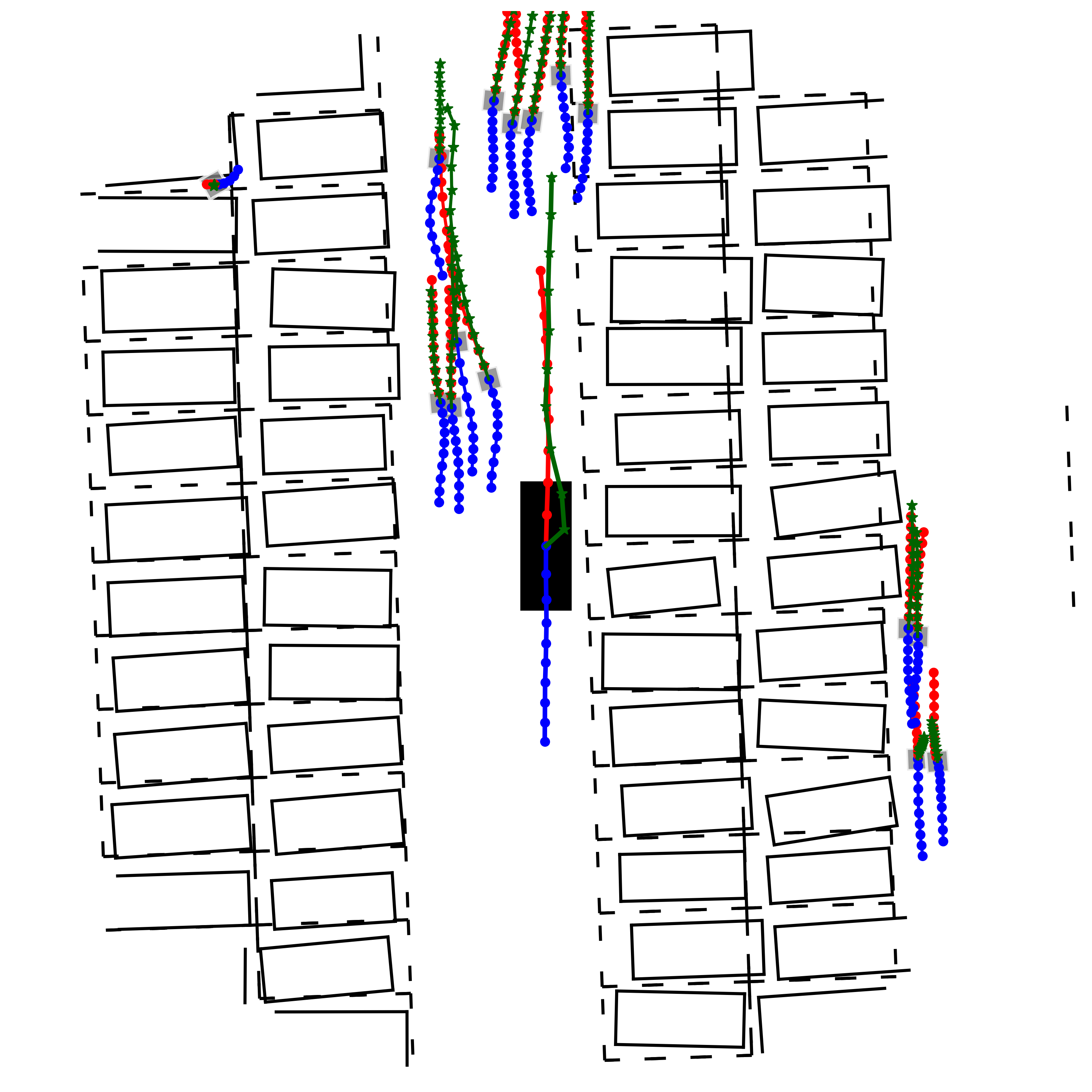}\\[0.3em]
    \end{minipage}
    \hfill
    \begin{minipage}[b]{0.24\textwidth}
        \centering
        \includegraphics[width=\linewidth]{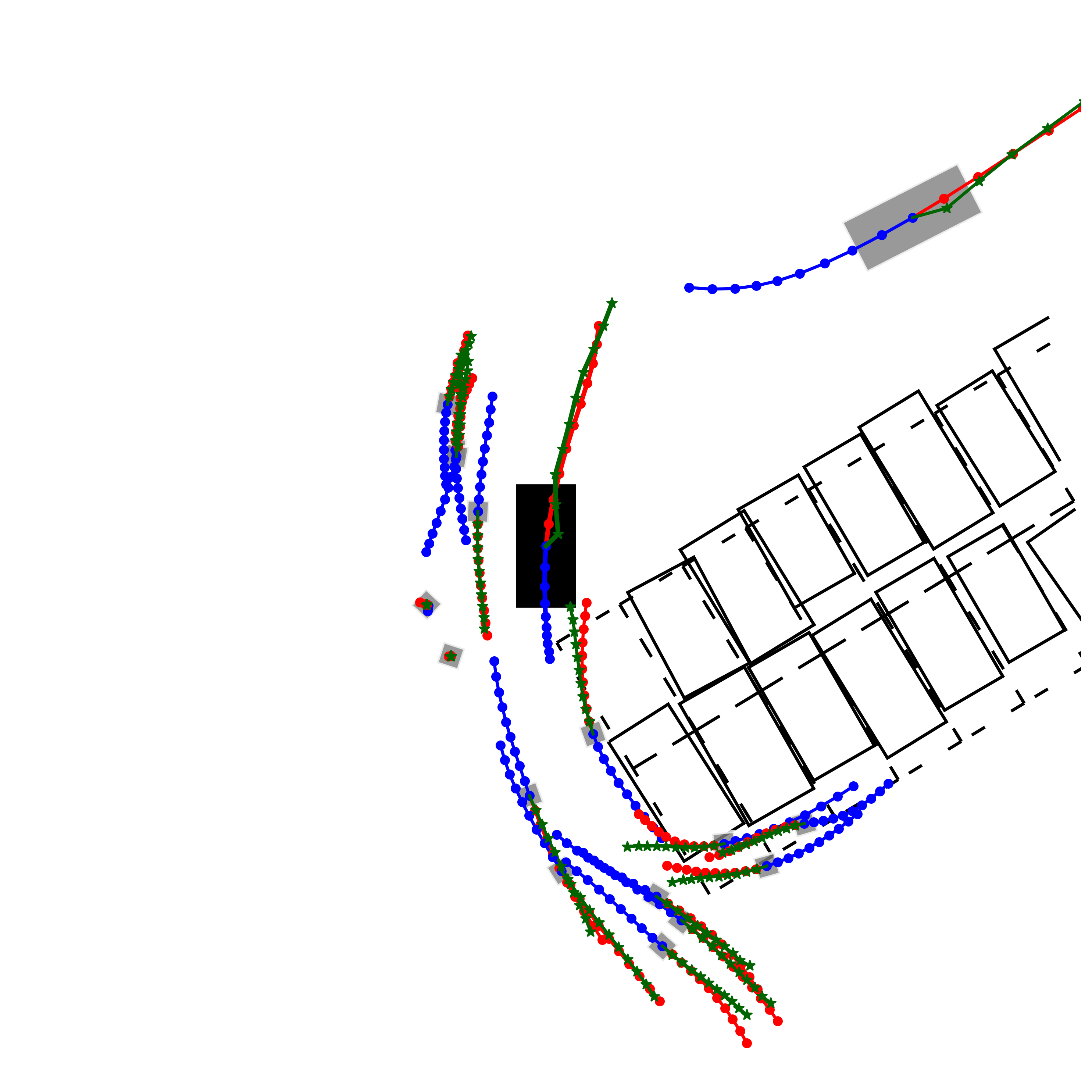}\\[0.3em]
    \end{minipage}
    
    \vspace{0.5cm}
    
    \begin{minipage}[b]{0.24\textwidth}
        \centering
        \includegraphics[width=\linewidth]{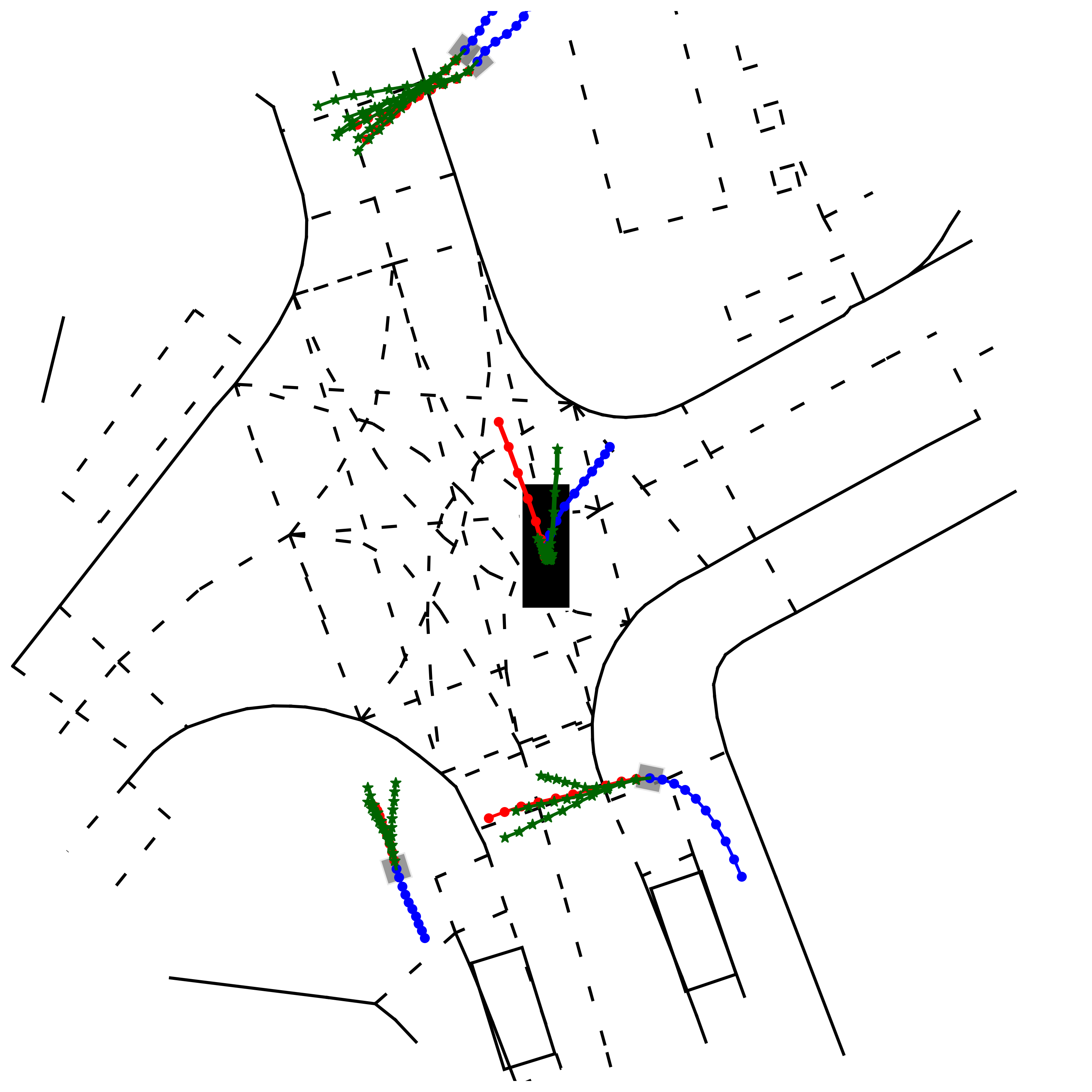}\\[0.3em]
    \end{minipage}
    \hfill
    \begin{minipage}[b]{0.24\textwidth}
        \centering
        \includegraphics[width=\linewidth]{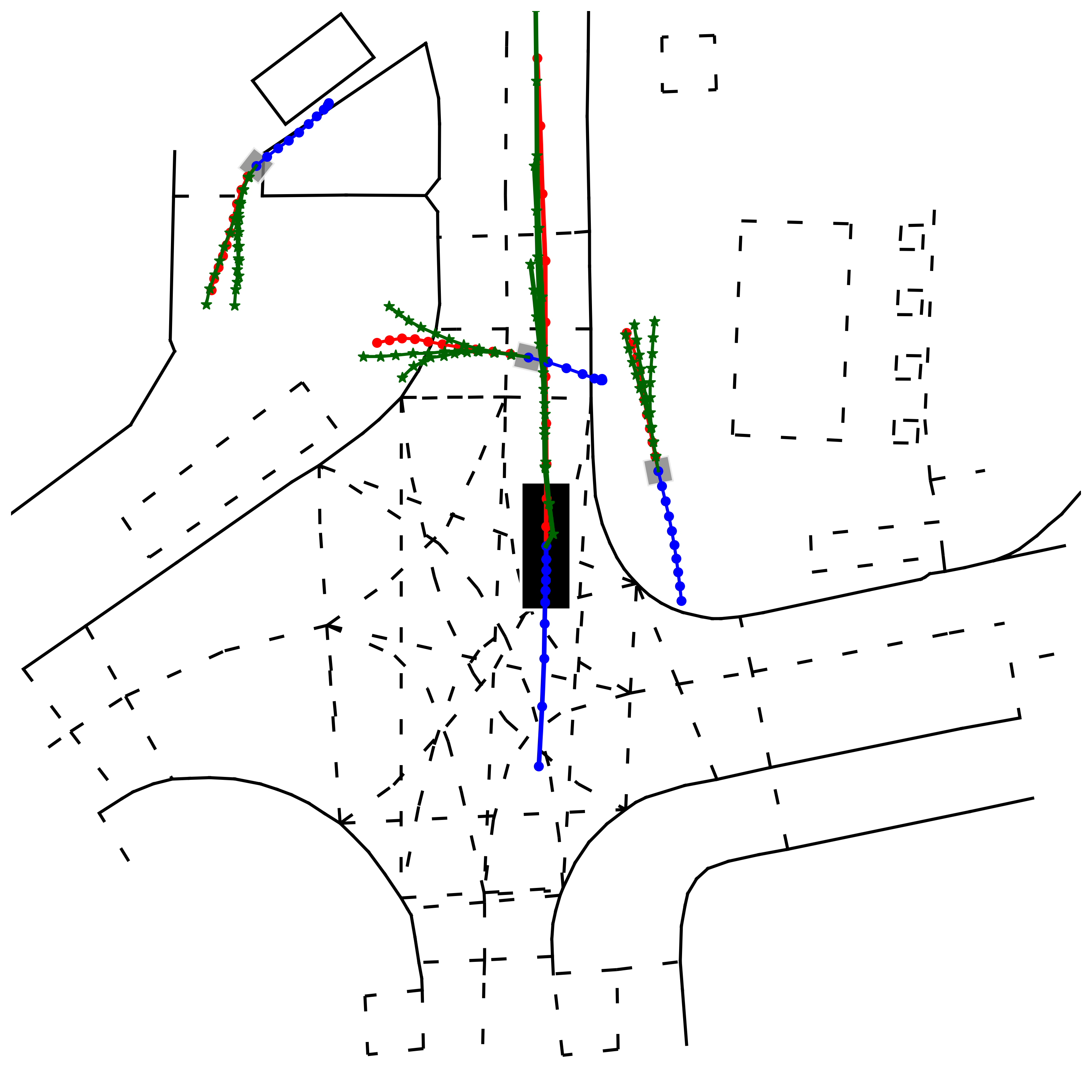}\\[0.3em]
    \end{minipage}
    \hfill
    \begin{minipage}[b]{0.24\textwidth}
        \centering
        \includegraphics[width=\linewidth]{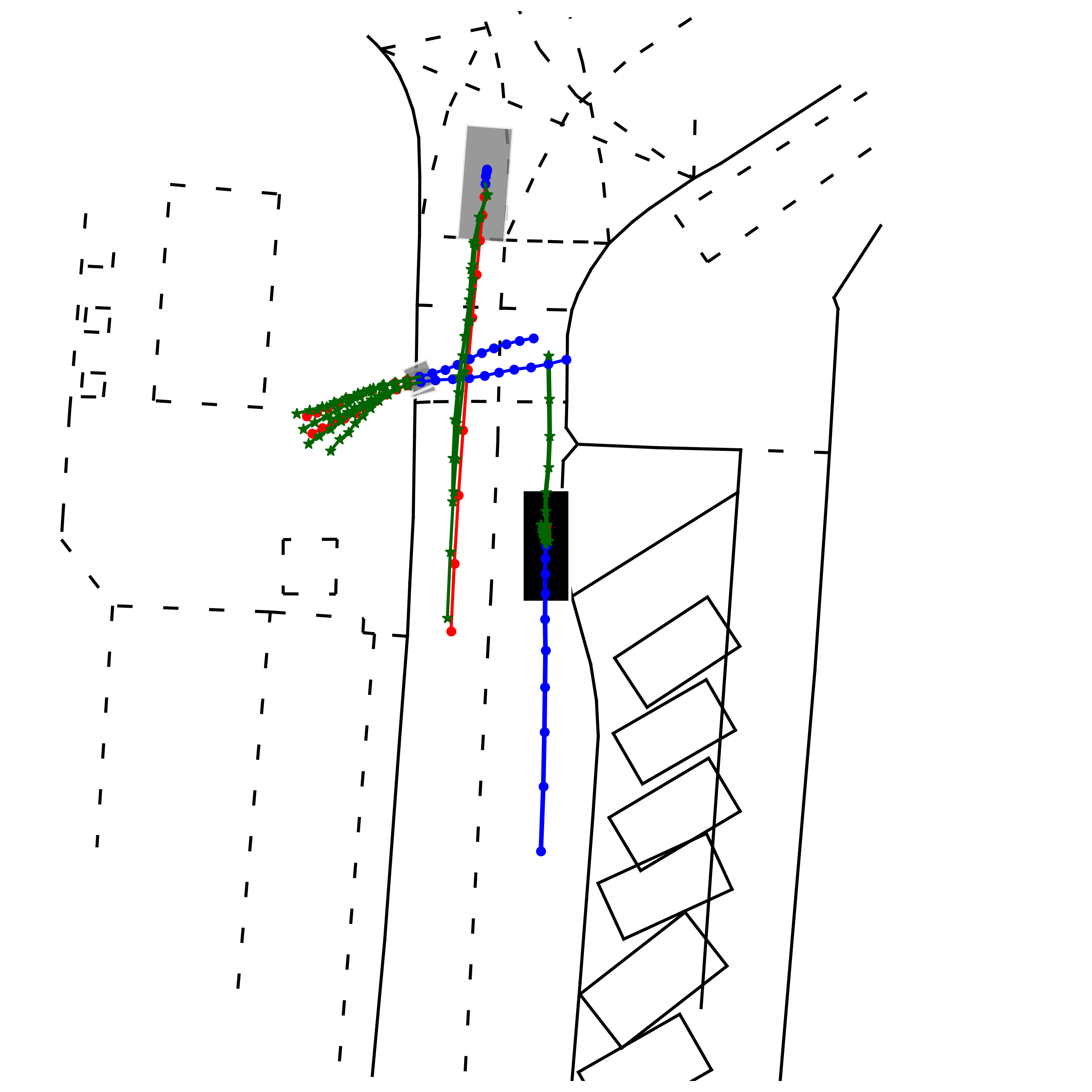}\\[0.3em]
    \end{minipage}
    \hfill
    \begin{minipage}[b]{0.24\textwidth}
        \centering
        \includegraphics[width=\linewidth]{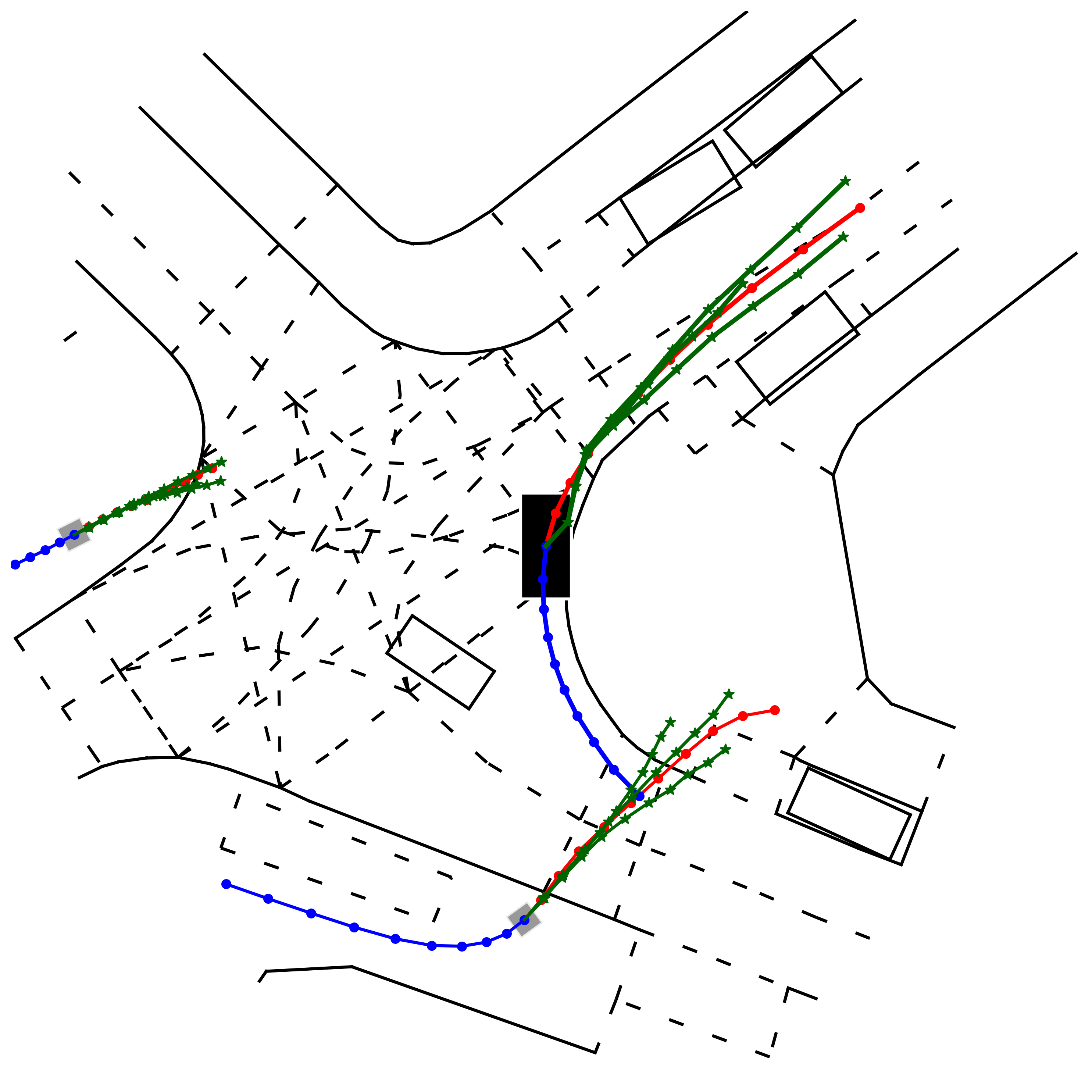}\\[0.3em]
    \end{minipage}
    
    \caption{Visualizations of \method predictions on the DLP (\textit{top row}) and inD (\textit{bottom row}) datasets. For clarity, on the DLP dataset with higher agent density, we show the trajectory with the highest probability from the multi-modal predictions; on the inD dataset with lower agent density, we present the top three most probable predictions to demonstrate the model's ability to capture multiple plausible futures. The ego vehicle is depicted in solid black at the center, while other agents are shown in gray. Trajectories are color-coded as follows: \textcolor{blue}{past} in blue, \textcolor{red}{ground-truth future} in red, and \textcolor{darkgreen}{predicted} in green.}
    \label{fig:main_figure}
    \vspace*{-.3cm}
\end{figure*}

\vspace*{-0.1cm}
\subsection{Setup and Implementation Details}
We train our model on the Dragon Lake Parking (DLP) dataset~\cite{shen2022parkpredict+} and the Intersections Drone (inD) dataset~\cite{bock2020ind}. 
Both datasets are preprocessed into ego-vehicle centric data samples, consisting of all the vehicles and pedestrians within a \qty{20}{\meter} range around the ego vehicle.
Our model predicts \(K = 6\) multi-modal trajectories of all agents in the data samples for the next \(T_f = 10\) timestamps, with a time interval of \qty{0.4}{\second}.
For comparison, we implement several trajectory prediction methods, including EKF, MultiPath++~\cite{varadarajan2022multipath++}, SceneTransformer~\cite{ngiam2021scene}, QCNet~\cite{zhou2023query}, and SIMPL~\cite{zhang2024simpl}.
We train all models using two NVIDIA RTX A6000 GPUs for 20 epochs to ensure convergence.

{\parskip=2pt
\noindent\textit{DLP Dataset:}
In the DLP dataset~\cite{shen2022parkpredict+}, we have agents of vehicle class (\textit{car}, \textit{medium vehicle}, \textit{bus}) and pedestrian class. 
Since ParkPredict+~\cite{shen2022parkpredict+} employs a sequential network that first predicts agent intentions from Bird's Eye View (BEV) images using a CNN and then feeds the features into a basic transformer, while \method uses polyline-based inputs, it is challenging to reproduce ParkPredict+ in our data preprocessing setting. Thus, we report the performance of the ParkPredict+ baseline based on their published results. For the other baselines, we integrate them into our data preprocessing pipeline and retrain them. Since the DLP dataset does not provide official data splits, we follow ParkPredict+, using 51,750 samples for training and 5,750 for validation.
}

{\parskip=2pt
\noindent\textit{inD Dataset:}
We also evaluate \method on the inD dataset by levelXdata~\cite{bock2020ind}. 
Although it is not an automated parking dataset, it includes intersection scenarios featuring a mix of vehicles and vulnerable road users (VRUs), such as pedestrians, rendering it highly relevant to our research focus.
The dataset consists of four traffic scenarios, from which we merge the ``Bendplatz'' and ``Frankenburg'' scenarios due to their high number of VRUs and then split the combined data into training and validation sets. 
After data preprocessing, we use 34,014 samples for training and 4,859 samples for validation.
}

\vspace*{-0.1cm}
\subsection{Baseline Comparison}

In this section, we compare \method with various baselines on the DLP and inD datasets.

{\parskip=3pt
\noindent\textit{DLP Dataset:}  
We report the performance on the DLP dataset in \cref{table:results-dlp}. We observe that \method substantially outperforms ParkPredict+, the only dedicated parking prediction baseline, as well as the driving prediction methods adapted to parking scenarios (EKF, MultiPath++, SceneTransformer, QCNet, and SIMPL). 
For vehicle trajectory prediction, \method achieves much lower minADE and minFDE scores compared to all other baselines. 
Most notably, our approach demonstrates strong performance in pedestrian prediction, with a significant margin of improvement relative to all other models, achieving a minADE of \qty{0.15}{\meter}, a minFDE of \qty{0.32}{\meter}, and an MR of \qty{0.6}{\percent}.
This enhanced accuracy is particularly critical in parking scenarios, where precise forecasting of pedestrian behavior is essential.
}

{\parskip=3pt
\noindent\textit{inD Dataset:} 
In \cref{table:results-ind}, we report the results on the inD dataset. 
For vehicle trajectory prediction, SIMPL achieves the lowest minADE and MR, while ParkDiffusion obtains the best minFDE. 
In pedestrian prediction, ParkDiffusion demonstrates a clear advantage, achieving a minADE of \qty{0.20}{\meter}, a minFDE of \qty{0.45}{\meter}, and an MR of \qty{1.8}{\percent}, compared to the minADE of MultiPath++~\cite{varadarajan2022multipath++} (\qty{0.43}{\meter}) and the minFDE and MR of SIMPL~\cite{zhang2024simpl} (\qty{0.49}{\meter}, \qty{2.2}{\percent}). 
Overall, our method achieves the best minADE and minFDE scores compared to all baselines and matches SIMPL for the lowest MR in the combined evaluation. 
These results demonstrate that our approach achieves robust performance across diverse traffic settings, showing strong generalization and consistent accuracy in vehicle and pedestrian trajectory predictions.
}

\vspace*{-0.1cm}
\subsection{Qualitative Results}
\cref{fig:main_figure} visualizes the prediction results of \method on both datasets. 
For the DLP dataset, our model accurately predicts the trajectories of vehicles and pedestrians, including turning and reverse maneuvers that are typically observed in automated parking scenarios. Moreover, the model maintains high prediction accuracy for pedestrians, even in environments crowded with multiple agents and complex interactions. In the inD dataset, we present multi-modal predictions for vehicles and pedestrians. These predictions effectively cover the ground-truth trajectory while capturing various possible movement patterns for different agents. The most challenging scenario is depicted in the bottom left figure, where a car is reversing in the intersection and switching to a new lane. Among the three candidate trajectories, one trajectory shows a slight deviation in the predicted direction while reflecting the vehicle's speed. In contrast, the other two trajectories correctly capture the directional change but underestimate the speed. Overall, the general behavior of reversing and initiating movement in a different lane is captured well. 

\vspace*{-0.1cm}
\subsection{Ablation Studies}
The ablation studies on the DLP dataset provide valuable insights into the impact of individual components and the robustness of our approach under varying conditions.


\begin{table}[!t]
\centering
\caption{Ablation Study: Components Analysis}
\vspace*{-0.3cm}
\label{tab:ablation-components}
\setlength\tabcolsep{1pt} 
\begin{threeparttable}
    \begin{tabular}{l|ccc|ccc|ccc}
        \toprule
         & \multicolumn{3}{c|}{\textbf{Vehicle}} & \multicolumn{3}{c|}{\textbf{Pedestrian}} & \multicolumn{3}{c}{\textbf{All}} \\
        \textbf{Method} & mADE & mFDE & MR & mADE & mFDE & MR & mADE & mFDE & MR \\
        \midrule
        Vanilla model 
        & 0.24 & 0.45 & 0.7
        & 0.26 & 0.52 & 1.5
        & 0.25 & 0.49 & 1.3 \\
        \hspace{1pt}+ Map encod.
        & 0.19 & 0.34 & 0.3
        & 0.18 & 0.37 & 0.9
        & 0.18 & 0.36 & 0.7 \\
        \hspace{1pt}+ Type embed.
        & 0.19 & 0.31 & \textbf{0.2}
        & 0.17 & 0.35 & 0.8
        & 0.18 & 0.34 & 0.6 \\
        \hspace{1pt}+ Kinematics
        & \textbf{0.17} & \textbf{0.24} & \textbf{0.2} 
        & \textbf{0.15} & \textbf{0.32} & \textbf{0.6} 
        & \textbf{0.16} & \textbf{0.30} & \textbf{0.5} \\
        \bottomrule
    \end{tabular}
    We analyze the impact of the individual steps of our proposed \method approach by iteratively integrating additional components. Note that each row includes all preceding components. We report the minADE (mADE) and minFDE (mFDE) in meters and the MR as a percentage. The best scores per metric are shown in bold.
\end{threeparttable}
\end{table}

\begin{table}[!t]
\centering
\caption{Ablation Study: Context Masking}
\vspace*{-0.3cm}
\label{tab:ablation-map-masking}
\setlength\tabcolsep{2pt} 
\begin{threeparttable}
    \begin{tabular}{c|ccc|ccc|ccc}
        \toprule
        \textbf{Masked} & \multicolumn{3}{c|}{\textbf{Vehicle}} & \multicolumn{3}{c|}{\textbf{Pedestrian}} & \multicolumn{3}{c}{\textbf{All}} \\
        \textbf{area} & mADE & mFDE & MR & mADE & mFDE & MR & mADE & mFDE & MR \\
        \midrule
        \qty{0}{\percent}   
        & \textbf{0.17} & \textbf{0.24} & \textbf{0.2} 
        & \textbf{0.15} & \textbf{0.32} & 0.6 
        & \textbf{0.16} & \textbf{0.30} & \textbf{0.5} \\
        \qty{25}{\percent}  
        & 0.18 & 0.26 & \textbf{0.2}
        & \textbf{0.15} & \textbf{0.32} & 0.7 
        & \textbf{0.16} & \textbf{0.30} & 0.6 \\
        \qty{50}{\percent}  
        & 0.19 & 0.27 & \textbf{0.2} 
        & 0.16 & 0.33 & 0.8 
        & 0.17 & 0.31 & 0.6 \\
        \qty{75}{\percent}  
        & 0.20 & 0.28 & 0.3 
        & 0.16 & 0.34 & 0.9 
        & 0.17 & 0.32 & 0.7 \\
        \qty{100}{\percent} 
        & 0.22 & 0.33 & 0.5 
        & 0.19 & 0.39 & 1.5 
        & 0.20 & 0.37 & 1.2\\
        \bottomrule
    \end{tabular}
    We assess the robustness of \method to imperfect map data by progressively and randomly masking input map polylines, starting from complete information (\qty{0}{\percent}) and increasing to no context data (\qty{100}{\percent}). We report the minADE (mADE) and minFDE (mFDE) in meters and the MR as a percentage. The best scores per metric are shown in bold.
\end{threeparttable}
\end{table}

\begin{table}[!t]
\centering
\caption{Ablation Study: Number of Agents}
\vspace*{-0.3cm}
\label{tab:ablation-agents}
\setlength\tabcolsep{1.4pt} 
\begin{threeparttable}
    \begin{tabular}{c|c|ccc|ccc|ccc}
        \toprule
        \textbf{Agent} & \textbf{Agent} & \multicolumn{3}{c|}{\textbf{Vehicle}} & \multicolumn{3}{c|}{\textbf{Pedestrian}} & \multicolumn{3}{c}{\textbf{All}} \\
        \textbf{count} & \textbf{ratio} & mADE & mFDE & MR & mADE & mFDE & MR & mADE & mFDE & MR \\
        \midrule
        \numrange{1}{4}   
        & 41.6 
        & \textbf{0.16} & \textbf{0.22} & 0.3 
        & 0.18 & 0.37 & 1.2 
        & \textbf{0.17} & 0.33 & 0.9 \\
        \numrange{5}{9}   
        & 33.4 
        & 0.24 & 0.34 & 0.3 
        & 0.17 & 0.36 & 1.0 
        & 0.19 & 0.35 & 0.8 \\
        \numrange{10}{14} 
        & 13.8 
        & 0.31 & 0.42 & 0.3 
        & \textbf{0.15} & 0.33 & \textbf{0.6}  
        & 0.20 & 0.36 & \textbf{0.5} \\
        \numrange{15}{19} 
        & 6.8 
        & 0.22 & 0.36 & \textbf{0.2} 
        & \textbf{0.15} & \textbf{0.31} & 0.7 
        & \textbf{0.17} & \textbf{0.32} & 0.6 \\
        \numrange{20}{24} 
        & 2.7 
        & 0.30 & 0.49 & 0.4 
        & \textbf{0.15} & 0.32 & \textbf{0.6}  
        & 0.19 & 0.37 & \textbf{0.5} \\
        $\geq 25$   
        & 1.8 
        & 0.34 & 0.59 & 0.3 
        & \textbf{0.15} & 0.32 & 0.7
        & 0.20 & 0.40 & 0.6\\
        \bottomrule
    \end{tabular}
    We assess the impact of scene complexity on model performance by segmenting the dataset according to the number of agents present in each sample. The agent ratio for each segment represents the percentage of samples within that range relative to the entire dataset. We report the minADE (mADE) and minFDE (mFDE) in meters, and both MR and the agent ratio as a percentage. The best scores per metric are shown in bold.
\end{threeparttable}
\end{table}


{\parskip=3pt
\noindent\textit{Components Analysis:}
In \cref{tab:ablation-components}, we present a detailed component-wise analysis of our approach.
The baseline model is implemented by incorporating the Leapfrog diffusion approach~\cite{mao2023leapfrog} with our heterogeneous multi-agent multi-modal trajectory prediction pipeline. 
Subsequently, we integrate the dual map encoder, agent type embedding, and kinematic constraints module into the baseline model step by step.
The results show that the map encoder substantially improves performance across the board. 
The agent type embeddings further refine the predictions, yielding a minor performance improvement. 
Finally, incorporating kinematic constraints further reduces prediction errors, achieving the best performance across all metrics.
}


{\parskip=3pt
\noindent\textit{Context Masking:}
In \cref{tab:ablation-map-masking}, we evaluate the impact of context masking on model performance. Starting with complete map information (\qty{0}{\percent} masking), the model achieves the best results for vehicles and pedestrians. As the amount of masking increases, thus limiting the available contextual data, the performance gradually degrades. 
Notably, while \qty{25}{\percent} masking results in only minor performance reductions, more extensive masking (\qty{50}{\percent} to \qty{100}{\percent}) leads to a marked increase in prediction errors. 
These results indicate that model performance deteriorates with the loss of contextual information, underscoring the importance of accurate map context to achieve reliable trajectory prediction.
}


{\parskip=3pt
\noindent\textit{Number of Agents:}
Finally, in \cref{tab:ablation-agents}, we study how the number of agents in a scene influences the performance of our proposed model. 
For this experiment, we categorized scenes according to their number of agents to analyze how an increase in agent density impacts the prediction accuracy. 
The results reveal that vehicle trajectory performance degrades as the number of agents increases, while pedestrian predictions tend to improve in denser scenes, potentially benefiting from richer interaction cues. 
Overall, although vehicle errors increase with higher agent counts, the improvements in pedestrian predictions lead to minor variations in the aggregated performance metrics across all agents.
}


\section{Conclusion} 
In this paper, we introduced \method, a novel diffusion-based framework for heterogeneous multi-agent trajectory prediction in automated parking scenarios. 
Our approach leverages several key components, including an agent encoder, a dual map encoder, an agent type embedding, and a kinematic refinement module. The overall goal of the approach is to capture complex environmental contexts and accurately forecast trajectories for both vehicles and pedestrians. 
Notably, \method demonstrates superior performance in pedestrian trajectory prediction, which is especially critical in traffic environments with a high prevalence of vulnerable road users. 
Extensive experiments, including baseline comparisons on two datasets and various ablation studies, validate the accuracy and robustness of our approach. 
In future work, we will investigate an extension of our model to additional agent types to further enhance the modeling of heterogeneous interactions.
\looseness=-1


\bibliographystyle{IEEEtran}
\bibliography{reference}

\begin{thebibliography}{10}
\providecommand{\url}[1]{#1}
\csname url@rmstyle\endcsname
\providecommand{\newblock}{\relax}
\providecommand{\bibinfo}[2]{#2}
\providecommand\BIBentrySTDinterwordspacing{\spaceskip=0pt\relax}
\providecommand\BIBentryALTinterwordstretchfactor{4}
\providecommand\BIBentryALTinterwordspacing{\spaceskip=\fontdimen2\font plus
\BIBentryALTinterwordstretchfactor\fontdimen3\font minus \fontdimen4\font\relax}
\providecommand\BIBforeignlanguage[2]{{%
\expandafter\ifx\csname l@#1\endcsname\relax
\typeout{** WARNING: IEEEtran.bst: No hyphenation pattern has been}%
\typeout{** loaded for the language `#1'. Using the pattern for}%
\typeout{** the default language instead.}%
\else
\language=\csname l@#1\endcsname
\fi
#2}}

\bibitem{mu2023inverse}
X.~Mu, H.~Ye, D.~Zhu, T.~Chen, and T.~Qin, ``Inverse perspective mapping-based neural occupancy grid map for visual parking,'' in \emph{{IEEE} Int. Conf. on Robotics and Automation}, 2023, pp. 8400--8406.

\bibitem{yang2021towards}
Y.~Yang, M.~Pan, S.~Jiang, J.~Wang, W.~Wang, J.~Wang, and M.~Wang, ``Towards autonomous parking using vision-only sensors,'' in \emph{{IEEE/RSJ} Int. Conf. on Intelligent Robots and Systems}, 2021, pp. 2038--2044.

\bibitem{liu2024less}
M.~Liu, X.~Tang, Y.~Qian, J.~Chen, and L.~Li, ``{LESS-map}: Lightweight and evolving semantic map in parking lots for long-term self-localization,'' in \emph{{IEEE} Int. Conf. on Robotics and Automation}, 2024.

\bibitem{kim2024visual}
J.~Kim, G.~Koo, H.~Park, and N.~Doh, ``Visual localization in repetitive and symmetric indoor parking lots using {3D} key text graph,'' in \emph{{IEEE} Int. Conf. on Robotics and Automation}, 2024, pp. 10\,185--10\,191.

\bibitem{dai2021long}
S.~Dai and Y.~Wang, ``Long-horizon motion planning for autonomous vehicle parking incorporating incomplete map information,'' in \emph{{IEEE} Int. Conf. on Robotics and Automation}, 2021, pp. 8135--8142.

\bibitem{zheng2024speeding}
X.~Zheng, X.~Zhang, and D.~Xu, ``Speeding up path planning via reinforcement learning in {MCTS} for automated parking,'' in \emph{{IEEE/RSJ} Int. Conf. on Intelligent Robots and Systems}, 2024, pp. 5410--5415.

\bibitem{zhou2023query}
Z.~Zhou, J.~Wang, Y.-H. Li, and Y.-K. Huang, ``Query-centric trajectory prediction,'' in \emph{{IEEE/CVF} Conf. on Computer Vision and Pattern Recognition}, 2023, pp. 17\,863--17\,873.

\bibitem{zhang2024simpl}
L.~Zhang, P.~Li, S.~Liu, and S.~Shen, ``{SIMPL}: A simple and efficient multi-agent motion prediction baseline for autonomous driving,'' \emph{{IEEE} Robotics and Automation Letters}, 2024.

\bibitem{distelzweig2024entropy}
A.~Distelzweig, A.~Look, E.~Kosman, F.~Janjo{\v{s}}, J.~Wagner, and A.~Valada, ``Entropy-based uncertainty modeling for trajectory prediction in autonomous driving,'' \emph{arXiv preprint arXiv:2410.01628}, 2024.

\bibitem{distelzweig2024motion}
A.~Distelzweig, E.~Kosman, A.~Look, F.~Janjo{\v{s}}, D.~K. Manivannan, and A.~Valada, ``Motion forecasting via model-based risk minimization,'' \emph{arXiv preprint arXiv:2409.10585}, 2024.

\bibitem{mao2023leapfrog}
W.~Mao, C.~Xu, Q.~Zhu, S.~Chen, and Y.~Wang, ``Leapfrog diffusion model for stochastic trajectory prediction,'' in \emph{{IEEE/CVF} Conf. on Computer Vision and Pattern Recognition}, 2023, pp. 5517--5526.

\bibitem{wong2024socialcircle}
C.~Wong, B.~Xia, Z.~Zou, Y.~Wang, and X.~You, ``{SocialCircle}: Learning the angle-based social interaction representation for pedestrian trajectory prediction,'' in \emph{{IEEE/CVF} Conf. on Computer Vision and Pattern Recognition}, 2024, pp. 19\,005--19\,015.

\bibitem{radwan2020multimodal}
N.~Radwan, W.~Burgard, and A.~Valada, ``Multimodal interaction-aware motion prediction for autonomous street crossing,'' \emph{{Int. Journal of Robotics Research}}, vol.~39, no.~13, pp. 1567--1598, 2020.

\bibitem{shen2022parkpredict+}
X.~Shen, M.~Lacayo, N.~Guggilla, and F.~Borrelli, ``{ParkPredict+}: Multimodal intent and motion prediction for vehicles in parking lots with {CNN} and transformer,'' in \emph{{IEEE} Int. Conf. on Intelligent Transportation Systems}, 2022, pp. 3999--4004.

\bibitem{bock2020ind}
J.~Bock, R.~Krajewski, T.~Moers, S.~Runde, L.~Vater, and L.~Eckstein, ``The {inD} dataset: A drone dataset of naturalistic road user trajectories at german intersections,'' in \emph{IEEE Intelligent Vehicles Symposium}, 2020.

\bibitem{ngiam2021scene}
J.~Ngiam, V.~Vasudevan, B.~Caine, \emph{et~al.}, ``{Scene Transformer}: A unified architecture for predicting future trajectories of multiple agents,'' in \emph{Int. Conf. on Learning Representations}, 2022.

\bibitem{varadarajan2022multipath++}
B.~Varadarajan, A.~Hefny, A.~Srivastava, \emph{et~al.}, ``Multipath++: Efficient information fusion and trajectory aggregation for behavior prediction,'' in \emph{{IEEE} Int. Conf. on Robotics and Automation}, 2022, pp. 7814--7821.

\bibitem{grimm2023heterogeneous}
D.~Grimm, M.~Zipfl, F.~Hertlein, A.~Naumann, J.~Luettin, S.~Thoma, S.~Schmid, L.~Halilaj, A.~Rettinger, and J.~M. Z{\"o}llner, ``Heterogeneous graph-based trajectory prediction using local map context and social interactions,'' in \emph{{IEEE} Int. Conf. on Intelligent Transportation Systems}, 2023, pp. 2901--2907.

\bibitem{fang2023heterogeneous}
J.~Fang, C.~Zhu, P.~Zhang, H.~Yu, and J.~Xue, ``Heterogeneous trajectory forecasting via risk and scene graph learning,'' \emph{{IEEE} Transactions on Intelligent Transportation Systems}, pp. 12\,078--12\,091, 2023.

\bibitem{chandra2019traphic}
R.~Chandra, U.~Bhattacharya, A.~Bera, and D.~Manocha, ``Traphic: Trajectory prediction in dense and heterogeneous traffic using weighted interactions,'' in \emph{{IEEE/CVF} Conf. on Computer Vision and Pattern Recognition}, 2019, pp. 8483--8492.

\bibitem{ma2019trafficpredict}
Y.~Ma, X.~Zhu, S.~Zhang, R.~Yang, W.~Wang, and D.~Manocha, ``{TrafficPredict}: Trajectory prediction for heterogeneous traffic-agents,'' in \emph{{AAAI} Conf. on Artificial Intelligence}, 2019.

\bibitem{li2023real}
L.~Li, X.~Wang, D.~Yang, Y.~Ju, Z.~Zhang, and J.~Lian, ``Real-time heterogeneous road-agents trajectory prediction using hierarchical convolutional networks and multi-task learning,'' \emph{{IEEE} Transactions on Intelligent Vehicles}, vol.~9, no.~2, pp. 4055--4069, 2023.

\bibitem{zhao2023trajectory}
S.~Zhao, M.~Li, T.~Huang, S.~Li, and Z.~Xing, ``Trajectory prediction for heterogeneous road-agents using dual attention model,'' \emph{Measurement}, vol. 212, p. 112685, 2023.

\bibitem{liu2024intention}
C.~Liu, S.~He, H.~Liu, and J.~Chen, ``Intention-aware denoising diffusion model for trajectory prediction,'' \emph{arXiv preprint arXiv:2403.09190}, 2024.

\bibitem{jiang2023motiondiffuser}
C.~Jiang, A.~Cornman, C.~Park, B.~Sapp, Y.~Zhou, D.~Anguelov, \emph{et~al.}, ``{MotionDiffuser}: Controllable multi-agent motion prediction using diffusion,'' in \emph{{IEEE/CVF} Conf. on Computer Vision and Pattern Recognition}, 2023, pp. 9644--9653.

\bibitem{li2023multi}
Z.~Li, H.~Liang, H.~Wang, X.~Zheng, J.~Wang, and P.~Zhou, ``A multi-modal vehicle trajectory prediction framework via conditional diffusion model: A coarse-to-fine approach,'' \emph{Knowledge-Based Systems}, vol. 280, p. 110990, 2023.

\bibitem{lv2024learning}
K.~Lv, L.~Yuan, and X.~Ni, ``Learning autoencoder diffusion models of pedestrian group relationships for multimodal trajectory prediction,'' \emph{{IEEE} Transactions on Instrumentation and Measurement}, 2024.

\bibitem{choi2024dice}
Y.~Choi, R.~C. Mercurius, S.~M.~A. Shabestary, and A.~Rasouli, ``{DICE}: Diverse diffusion model with scoring for trajectory prediction,'' in \emph{IEEE Intelligent Vehicles Symposium}, 2024, pp. 3023--3029.

\bibitem{zhu2024controltraj}
Y.~Zhu, J.~J. Yu, X.~Zhao, Q.~Liu, Y.~Ye, W.~Chen, Z.~Zhang, X.~Wei, and Y.~Liang, ``{ControlTraj}: Controllable trajectory generation with topology-constrained diffusion model,'' in \emph{{ACM SIGKDD} Conf. on Knowledge Discovery and Data Mining}, 2024, pp. 4676--4687.

\bibitem{li2023bcdiff}
R.~Li, C.~Li, D.~Ren, G.~Chen, Y.~Yuan, and G.~Wang, ``{BCDiff}: Bidirectional consistent diffusion for instantaneous trajectory prediction,'' \emph{Conf. on Neural Information Processing Systems}, 2023.

\bibitem{li2023avm}
Y.~Li, W.~Yang, D.~Lin, Q.~Wang, Z.~Cui, and X.~Qin, ``{AVM-SLAM}: Semantic visual {SLAM} with multi-sensor fusion in a bird's eye view for automated valet parking,'' in \emph{{IEEE/RSJ} Int. Conf. on Intelligent Robots and Systems}, 2023.

\bibitem{kang2021robust}
Y.~Kang, Y.~Song, W.~Ge, and T.~Ling, ``Robust multi-camera {SLAM} with {Manhattan} constraint toward automated valet parking,'' in \emph{{IEEE/RSJ} Int. Conf. on Intelligent Robots and Systems}, 2021.

\bibitem{leu2022autonomous}
J.~Leu, Y.~Wang, M.~Tomizuka, and S.~Di~Cairano, ``Autonomous vehicle parking in dynamic environments: An integrated system with prediction and motion planning,'' in \emph{{IEEE} Int. Conf. on Robotics and Automation}, 2022, pp. 10\,890--10\,897.

\bibitem{li2024parkinge2e}
C.~Li, Z.~Ji, Z.~Chen, T.~Qin, and M.~Yang, ``{ParkingE2E}: Camera-based end-to-end parking network, from images to planning,'' in \emph{{IEEE/RSJ} Int. Conf. on Intelligent Robots and Systems}, 2024, pp. 13\,206--13\,212.

\bibitem{shen2020parkpredict}
X.~Shen, I.~Batkovic, V.~Govindarajan, P.~Falcone, T.~Darrell, and F.~Borrelli, ``{ParkPredict}: Motion and intent prediction of vehicles in parking lots,'' in \emph{IEEE Intelligent Vehicles Symposium}, 2020.

\bibitem{cui2020deep}
H.~Cui, T.~Nguyen, F.-C. Chou, T.-H. Lin, J.~Schneider, D.~Bradley, and N.~Djuric, ``Deep kinematic models for kinematically feasible vehicle trajectory predictions,'' in \emph{{IEEE} Int. Conf. on Robotics and Automation}, 2020, pp. 10\,563--10\,569.

\bibitem{westny2024diffusion}
T.~Westny, B.~Olofsson, and E.~Frisk, ``Diffusion-based environment-aware trajectory prediction,'' \emph{arXiv preprint arXiv:2403.11643}, 2024.

\end{thebibliography}


\end{document}